\documentclass[aps,prl,preprint,groupedaddress]{revtex4-1}

\usepackage{subfigure}
\usepackage{lineno}
\usepackage{hyperref}
\usepackage{multirow}
\usepackage{array,graphicx}
\usepackage{rotating}
\usepackage{booktabs}
\usepackage{rotating}

\begin{document}

\title{Authorship Attribution Based on Life-Like Network Automata}%

%\author{The American Physical Society}%
%\email[REVTeX Support: ]{revtex@aps.org}
%\affiliation{1 Research Road, Ridge, NY 11961}
%\date{August 10, 2010}%

\author{Jeaneth Machicao$^1$}
\author{Edilson A. Corr\^{e}a Jr.$^2$}
\author{Gisele H. B. Miranda$^2$}
\author{Diego R. Amancio$^2$}
\author{Odemir M. Bruno$^{1,2}$}

\affiliation{1 S\~{a}o Carlos Institute of Physics, University of S\~{a}o Paulo, S\~{a}o Carlos - SP, PO Box 369, 13560-970, Brazil.\\
2 Institute of Mathematics and Computer Science, University of S\~{a}o Paulo, S\~{a}o Carlos - SP, 13560-970, Brazil.}
%\affil[*]{bruno@ifsc.usp.br}

\begin{abstract}
The authorship attribution is a problem of considerable practical and technical interest. Several methods have been designed to infer the authorship of disputed documents in multiple contexts. While traditional statistical methods based solely on word counts and related measurements have provided a simple, yet effective solution in particular cases; they are prone to manipulation. Recently, texts have been successfully modeled as networks, where words are represented by nodes linked according to textual similarity measurements. Such models are useful to identify informative topological patterns for the authorship recognition task. However, there is no consensus on which measurements should be used. Thus, we proposed a novel method to characterize text networks, by considering both topological and dynamical aspects of networks. Using concepts and methods from cellular automata theory, we devised a strategy to grasp informative spatio-temporal patterns from this model. Our experiments revealed an outperformance over traditional analysis relying only on topological measurements. Remarkably, we have found a dependence of pre-processing steps (such as the lemmatization) on the obtained results, a feature that has mostly been disregarded in related works. The optimized results obtained here pave the way for a better characterization of textual networks.
\end{abstract}

\maketitle
%\tableofcontents

\section{Introduction}

The current massive production of data has brought up plenty of challenges to the areas of Data Mining, Natural Language Processing (NLP) and Machine Learning. An example of a current challenge in information sciences is the authorship attribution task, which amounts to the ability to assign authorship to anonymous or disputed documents. This task has drawn attention from researchers mostly for its implications in real applications, such as plagiarism detection~\cite{FrancoSalvador2016550,Labbe}, forensics against cyber crimes~\cite{Vacca:2005:CFC:1076307} and resolution of disputed documents~\cite{ASI:ASI21001}.

Several methods have been proposed to undertake the authorship attribution problem~\cite{ASI:ASI21001}. Traditional techniques use text analytics and natural language processing concepts to characterize authors' writing styles~\cite{ASI:ASI21001}. For example, in several studies, it has been shown that the raw frequency of function words or the intermittency of content words is notably useful to discriminate authors' styles~\cite{amancio2015authorship,Brennan:2012:ASC:2382448.2382450}. In recent years, deeper paradigms have been employed to tackle this problem. Syntactical and semantical features are some examples of features not relying only on simple statistical analyses~\cite{Halteren:2007:AVL:1187415.1187416}. Despite being effective in particular contexts, deeper paradigms require a more complex data handling, a painstaking effort that may not yield good results in generic scenarios. Even though methods based on simple statistical analyses yield, in general, excellent results with the advantage of not requiring a large corpora for training or language-dependent resources, they are prone to manipulation via obfuscation of imitation attacks~\cite{Brennan:2012:ASC:2382448.2382450}. For this reason, more robust statistical methods have been proposed.

A recent trend in authorship attribution research is using the complex network framework, due to the success of its use in related tasks, mostly in text classification tasks~\cite{Martinici,Dorogovtsev2603,0295-5075-100-5-58002,10.1371/journal.pone.0067310,0295-5075-93-2-28005,0295-5075-83-1-18002,Mehri20122429}. In this paradigm, documents are modeled by means of a co-occurrence network~\cite{10.1371/journal.pone.0067310}, and the properties of the formed networks are used as authors' fingerprints in the classification process~\cite{amancio2011comparing}. Although such methods have proven useful for discriminating writing styles with a certain robustness provided by topological analysis, they usually provide no better results than traditional techniques based e.g. in n-grams models when used as a single source of text characterization. However, complex network topologies are less prone to manipulation, which makes these network-based methods more robust in real scenarios. Note that complex network-based measurements provide a complementary view of unstructured documents, a feature that can be further explored in hybrid approaches.

In a typical networked-based authorship recognition system, texts are modeled as a network and the structure of these networks is then used as a relevant feature to discriminate distinct authors~\cite{amancio2011comparing}. While traditional network topology measurements are useful to understand the main topological properties of texts, they may provide an ambiguous characterization, mainly when subtleties in style are not mapped into equivalent informative network structures. For this reason, the creation of informative, efficient and unambiguous network measurements for specific models remains as an open problem in network science. In this context, we explore a novel network characterization based on cellular automata theory (CA)~\cite{WOLFRAM19841}.

In the last decade, the fusion of networks and cellular automata, appeared into the Literature~\cite{watts1999small,tomassini2005evolution,marr2012cellular,lifelikeNA}. This discrete dynamical system, called as Network Automata (NA), uses the network structure as the tessellation of the cellular automaton, whose dynamics is governed by a rule that defines the states of its nodes at each time step. NAs turned out to be a powerful tool for pattern recognition purposes because it combines the advantages of the networks for modeling and analyses with the capabilities of CAs to extract complex patterns~\cite{gonccalves2012complex,lifelikeNA}.

In this manuscript, we propose a method to characterize networks representing written texts to tackle the authorship attribution task. The proposed method is based on Life-Like Network Automata (LLNA)~\cite{lifelikeNA}, which was inspired by the 2D Life-like CA~\cite{GardnerMathematicalGT}, a well-known set of rules explored in diverse fields~\cite{Soto2015TheXU,machicao2012chaotic,Broderick2004ALV,CsuhajVarj1997EcoGrammarSA}. We depart from the well-known word-adjacency model and include a LLNA dynamics to characterize text networks.
More specifically, our approach relies on a selection of informative LLNA's rules and, therefore, we expect to obtain spatio-temporal patterns possessing two important properties: (i) the books written by the same author displays similar patterns; and (ii) books written by distinct authors display distinct spatio-temporal patterns. Using a collection of texts written by 8 authors, we obtained an accuracy of 76\%, which is considerably more accurate than traditional methods based solely on topological properties of networks and, therefore, demonstrating the good performance of the proposed method.

\section{Material and methods}
\label{sec:material}

\subsection{Proposal overview}

\begin{figure}[h!]
	\centering
	\includegraphics[scale=0.65]{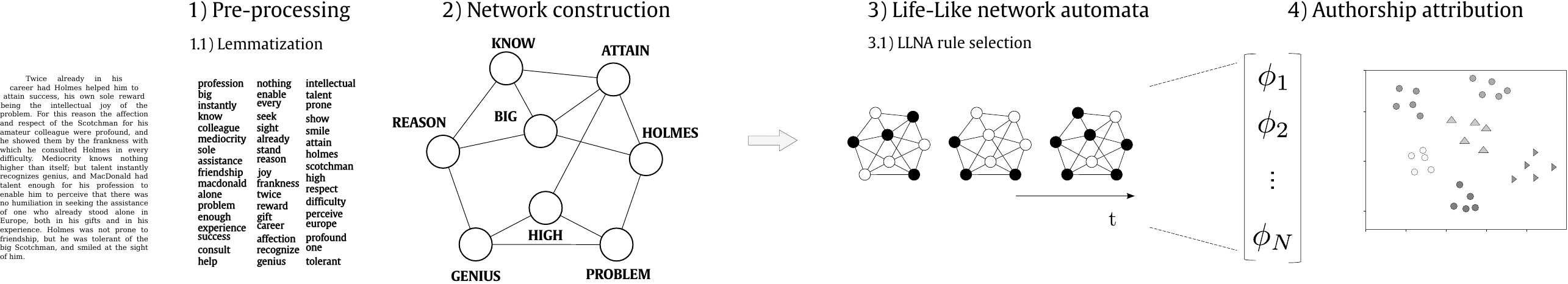}
 \caption{Authorship attribution framework based on LLNA method. The following steps are applied: (1) a written text is pre-processed; (2) a network is generated based on the extraction of keywords from the pre-processing; (3) a selected LLNA rule evolves over the textual network topology; (4) spatio-temporal features from the LLNA are extracted and then are used for the authorship attribution task.}
	\label{fig:procedure}
\end{figure}

In this section, we introduce an overview of the main proposal (see Figure~\ref{fig:procedure}) to understand not only the sequence of mathematical preliminaries, but also the experiments setup that are presented in Section~\ref{sec:results}. First, we introduce the well-known network model of text representation, the word-adjacency model. We also present optional text pre-processing strategies which may be applied to improve the characterization of texts. Some network measurements used to explore the properties of networks are presented. Next, we discuss the Life-Like network automata representation used in this article and their respective measurements.
The measurements extracted from the Life-Like network automata dynamics are then used to characterize the style of each author.

\subsection{Modeling and characterizing texts as networks}
\label{method:modeling}
In recent years, distinct ways to model texts as complex networks and graphs have been proposed~\cite{mihalcea2011graph}. Particularly, in the current study, we have used the so-called word adjacency (or co-occurrence) model, as it has been proven useful to grasp stylistic textual patterns~\cite{sole2010language,amancio2011using,amancio2011comparing}. In this model, each node represents a word and the edges are created whenever two words appear as adjacent in the raw text. Mathematically, the word adjacency network is represented by an adjacency matrix $A$, whose elements $A_{ij}$ are defined as
\begin{equation} \label{lab.tiebreaker}
 A_{ij} =\left\{
 \begin{array}{l l}
 1, & \textrm{if $i$ and $j$ are connected, and} \\
 0, & \textrm{otherwise.}
\end{array}\right.
\end{equation}

\subsubsection{Network construction}

Prior to the transformation of the text as a network, some pre-processing steps may be required. In most of the applications devoted to represent texts as networks, the three following steps are performed. The first step is the \emph{tokenization}, which is responsible to split the document into meaningful units, such as words and punctuation marks. The second step performs the removal of \emph{stopwords}, which are the words conveying little semantic meaning such as articles and prepositions. The list of stopwords is shown in {Section S3 of the Supplementary Information\footnote{The Supplementary Information is available at \url{https://dl.dropboxusercontent.com/u/2740286/automata.pdf}.}.} Note that, in this phase, punctuation marks are also disregarded, as they do not contribute to the semantic meaning of text. Finally, the third step, a lemmatization is applied to map the remaining words into their canonical forms. As such, verbs and nouns are mapped to their infinitive and singular forms, respectively. The lemmatization process usually requires the identification of the individual parts-of-speech to solve possible ambiguities. In this paper, we have used the Average Perceptron part-of-speech Tagger proposed by Collins~\cite{collins2002discriminative}. An exemplification of the pre-processing steps of a text extracted from the book \textit{The Valley of Fear} by Doyle, is shown in {Section S4 of the Supplementary Information}. %The obtained network of the mentioned text is illustrated in Figure \ref{fig:rede}.
%
%\begin{figure}[h!]
%	\centering
%	\includegraphics[scale=0.5]{2-eps-converted-to.pdf}
%    %\includegraphics[scale=0.5]{2}
% \caption{Exemplification of the network modeling using a short text extracted from \textit{The Valley of Fear} by Arthur Conan Doyle. In this example, we considered the lemmatization of all words to construct the network.}
%	\label{fig:rede}
%\end{figure}

Although lemmatization is often used in NLP tasks, Toman~\cite{toman2006influence} argued that this pre-processing step does not affect the performance of general text classification systems. To our knowledge, there is no systematic analysis on the effect of lemmatization on network-based authorship recognition methods. For this reason, we have considered the following three variations in the application of pre-processing in raw texts:
(i) \textit{none}, no lemmatization is performed; (ii) \textit{partial}, only nouns are lemmatized; and (iii) \textit{full}, all words are lemmatized, as it is done in more traditional works.

\subsubsection{Network measurements}
\label{Method:networkmeasurements}

In this section, we present a brief description of measurements used to characterize the topological properties of complex networks. These measurements are used here to study how the properties of text networks vary with distinct pre-processing steps. In addition, these measurements are also used for comparison and validation purposes.

The simplest measurements are the number of nodes ($N$) and edges ($E$). The density of a network is defined as $d = E / N(N-1)$, i.e. the fraction between the total number of edges and the maximum possible number of edges obtained in an equivalent fully connected network.

The degree $k_i$ of a node $i$ is defined as the number of neighbors that $i$ and is given by
\begin{equation}
	k_i=\sum_{j=1}^{N} A_{ij}.
\end{equation}	
The coefficient $\gamma$ of the degree distribution $P(k)=k^{-\gamma}$ is another widely known measurement in network science~\cite{Newman:2010:NI:1809753}. Similar to other real-world networks, text adjacency networks display the scale-free behavior~\cite{10.1371/journal.pone.0067310}. To estimate the coefficient $\gamma$, we used the strategy defined in~\cite{Clauset-PowerLawMatlab}. The degree is also usually measured in global terms as
\begin{equation} \label{avgdegree}
	\langle k \rangle = \frac{1}{N} \sum_{i=1}^{N} \sum_{j=1}^{N} A_{ij}.
\end{equation}	
The quantity defined in equation \ref{avgdegree} is the average degree, a measurement that has been applied in a myriad of network contexts~\cite{Newman:2010:NI:1809753}, even though many of the studied distributions makes this quantity not a representative element of the distribution, as many networks display a fat-tailed behavior~\cite{Li2016649,10.1371/journal.pone.0110121,sscoor,0295-5075-99-2-28002}. This is the case of text networks, whose fat-tailed degree distribution stems from the Zipf's law~\cite{lantiq}. However, in several cases, the average degree is useful to discriminate distinct topologies~\cite{Newman:2010:NI:1809753}.

Another well known connectivity measurement is the hierarchical degree $k^{h}$, which corresponds to the number of neighbors at distance $h$. This is a simple extension of the concept of node degree for further hierarchies. Despite its seeming simplicity, the use of hierarchies has proven useful to improve the characterization of several real-world networks~\cite{amancio2011using}.

While the degree is essentially a local measurement, some other indexes were specially devised to characterize the global topology of networks. This is the case of distance-based metrics. Measurements based on geodesic paths include the average shortest path length ($\langle L \rangle$) and the diameter ($D$). The average shortest path length of a network is computed as
\begin{equation}
	\langle L \rangle= \sum_{i=1}^N \sum_{j=1}^N \frac{d_{ij}}{N(N-1)},
\end{equation}
where $d_{ij}$ is the length of the shortest distance between nodes $i$ and $j$. The diameter of a network $D$, is the largest path length among all distances.

The transitivity of the network was measured by the average clustering coefficient $\langle C \rangle = 1/N \sum cc_i$, where $cc_i$ is the clustering coefficient computed for node $i$ and measure the probability of any two neighbors of $i$ being linked. Mathematically, the local clustering coefficient is computed as
\begin{equation}
	cc_i = \frac{2e_i}{k_i(k_i-1)},
\end{equation}
where $e_i$ represents the number of edges between the neighbors of node $i$. Even though this measure was originally used in social sciences, the clustering coefficient has been used to identify the specificity of words in distinct contexts.

Finally, we used the assortativity measure to measure if similar nodes are connected to each other. In this case, we used the concept of degree correlation, which assigns a high assortativity value for networks with edges established mostly between nodes with similar degree~\cite{newman2002}. The assortativity is given by
\begin{equation}
	\Gamma= \frac{(1/E)\sum_{j>i} k_ik_jA_{ij} - [(1/E)\sum_{j>i} (1/2)(k_i+k_j)A_{ij}]^2}{(1/E)\sum_{j>i} (1/2)(k_i^2+k_j^2)A_{ij} - [(1/E)\sum_{j>i} (1/2)(k_i+k_j)A_{ij}]^2}\,.
\label{Eq:assortativity}
\end{equation}
In general, text networks are disassortative, i.e. $\Gamma<0$~\cite{10.1371/journal.pone.0067310}.

\subsection{Life-Like network automata}
\label{Method:LifeLikeCA}

A network automata can be defined as a tuple $\mathcal{C}= \langle \mathcal{T},\mathcal{S},s, s_0,\phi\rangle$. $\mathcal{T}$ represents the NA space, which is the topology of a network comprising $N$ nodes (cells). $\mathcal{S}$ is the set of binary states $s_i$, where $s_i=1$ is the live state and $s_i=0$ the dead state. The cell's state can be identified by the function $s$, such that $s(c_i,t)$ gives the state of cell $c_i$ at time $t$. Finally, $s_0$ represents the initial configuration of all cells (i.e. the configuration at $t=0$) and $\phi$ is a transition function, i.e., the rule that governs the NA dynamics by defining how cells states are updated over time ~\cite{lifelikeNA}. Hereafter, we consider that the automata dynamics is stopped when $t=T$.

The LLNA was proposed as a class of binary NA inspired by the rules of the Life-like Cellular Automata (CA)~\cite{lifelikeNA}, which uses a set of outer-totalistic rules, \textit{i.e.,} rules that depend on the current state of cell $c_i$ and on the states of its neighboring cells. The LLNA transition function $\phi$ is stated as
\begin{equation} \label{eq:rule}
s(c_i, t+1)=\left\{
 \begin{array}{l l}
 1, & \textrm{if } s(c_i,t)=0~\textrm{and}~{x}/{r}\leq \rho_i< {(x+1)}/{r} \Rightarrow \textrm{born (B) rule} \\
 1, & \textrm{if } s(c_i,t)=1~\textrm{and}~{y}/{r}\leq \rho_i< {(y+1)}/{r} \Rightarrow \textrm{survive (S) rule} \\
 0, & \textrm{otherwise},
\end{array}\right.
\end{equation}

%\begin{equation} \label{eq:rule}
% s(c_i, t+1)=
%\begin{cases}
% 1, & \text{if } s(c_i,t)=0~\textrm{and}~{x}/{r}\leq \rho_i< {(x+1)}/{r} \Rightarrow \textrm{born (B) rule} \\
% 1, & \text{if } s(c_i,t)=1~\textrm{and}~{y}/{r}\leq \rho_i< {(y+1)}/{r} \Rightarrow \textrm{survive (S) rule} \\
% 0, & \text{otherwise},
%\end{cases}
%\end{equation}
%
where the neighborhood density $\rho_i$ of node $i$ is the proportion of alive neighbors, i.e.
\begin{equation}
	\rho_i=\frac{1}{k_i}\sum_{j=1}^{N} A_{ij}s(c_j,t).
\end{equation}
In the LLNA method, $r=9$ due to Moore's neighborhood~\cite{lifelikeNA}. As a consequence, there exists a total of $2^{18}$ possible transition rules in the Life-Like family of rules~\cite{lifelikeNA}.
In equation \ref{eq:rule}, the parameters $x$ and $y$ serve to label the rule in the form B$x_0x_1\ldots x_8$-S$y_0y_1\ldots y_8$, where B and S stand for ``born'' and ``survive'', respectively; and $x_i$ and $y_i$ are the possible $r=9$ digits in the rule described by equation \ref{eq:rule}. For instance, the rule B3-S23 is given by
\begin{equation} \label{eq:ruleex}
 s(c_i, t+1)=\left\{
 \begin{array}{l l}
1,& \text{if } s(c_i,t)=0~\textrm{and}~{3}/{r}\leq \rho_i< {4}/{r}\\
 1,& \text{if } s(c_i,t)=1~\textrm{and}~({2}/{r}\leq \rho_i< {3}/{r}~\textrm{or}~{3}/{r}\leq \rho_i< {4}/{r})\\
 0,   & \text{otherwise}.
\end{array}\right.
\end{equation}

%\begin{equation} \label{eq:ruleex}
% s(c_i, t+1)=
%\begin{cases}
% 1,& \text{if } s(c_i,t)=0~\textrm{and}~{3}/{r}\leq \rho_i< {4}/{r}\\
% 1,& \text{if } s(c_i,t)=1~\textrm{and}~({2}/{r}\leq \rho_i< {3}/{r}~\textrm{or}~{3}/{r}\leq \rho_i< {4}/{r})\\
% 0,   & \text{otherwise}.
%\end{cases}
%\end{equation}
%

\subsubsection{LLNA measurements}
\label{Method:LLNAmeasurements}

The dynamic of a network automata provides a global spatio-temporal pattern of evolution. Thus, each network node can be analyzed as a sequence of ones and zeros. A set of measurements, such as the Shannon entropy and Lempel-Ziv complexity were suggested to extract quantitative properties from the generated spatio-temporal patterns~\cite{lifelikeNA}.

The Shannon entropy of a binary sequence is defined as
\begin{equation}
{\mu_S}_i = -(p_i^0 \log_2{p_i^0} + p_i^1 \log_2{p_i^1}),
\end{equation}
where $p_i^1$ and $p_i^0$ are the probability of having ones and zeros in the sequence, respectively~\cite{shannon19481mathematical}. The Shannon entropy ranges in the interval $[0,1]$, where oscillating and complex spatio-temporal patterns tend to higher entropy values, while steady patterns tend to lower values.

The Lempel-Ziv complexity ${\mu_L}_i$, different from Shannon entropy, is a measurement based on the number of different blocks ($g$) that a sequence can contain~\cite{LEMPELZIV76}.
A minimum block is defined using the first bit on the left of the sequence. Then, one moves rightward, bit by bit, until an unseen subsequence appears, which is formed starting exactly after a previous block and ending at the current position. For instance, the binary sequence $11110001000111010010$ of length $l=20$, can be divided into $g=9$ {minimum blocks}: $1\vert 11\vert 10\vert 0\vert 01\vert 00\vert 011\vert 101\vert 001$. Given the number of blocks $g$, the Lempel-Ziv complexity is computed as
\begin{equation}
	{\mu_L}_i = \frac{g\log {l}} {l}.
\end{equation}
In literature, there exist several statistical similarity measurements designed to compare two binary sequences $p$ and $q$~\cite{Lesot:2009:SMB:1479242.1479248}. Most of these measurements are defined in terms of the following binary instances $a$=$pq$, $b=\bar{p}q$, $c=p \bar{q}$ and $d=\bar{p} \bar{q}$. The most traditional measurements are
\begin{equation}
 \textrm{3W-Jaccard:~}\frac{3a}{3a+b+c},~
 \textrm{Sokal \& Michener:~}\frac{a+d}{a+b+c+d},~\textrm{and}~
 \textrm{Sokal \& Sneat:~} \frac{a}{a+2b+2c}. \nonumber
\end{equation}

In our experiments, we have compared binary sequences by considering both spatial and temporal patterns. If we consider the spatial pattern, binary sequences generated by all nodes in two distinct time steps are compared. Analogously, if one considers temporal patterns, sequences generated by two nodes are compared by considering all times. In short, the spatio-temporal states of nodes can be represented in a matrix form, whose element stored in the $i$-th row and $t$-th column represents the state of node $i$ at time $t$. Thus, spatial patterns are analyzed via comparison of horizontal sequences, while temporal patterns are analyzed by comparing vertical sequences. Let $p_t$ be a horizontal sequence obtained at the $t$-th time step and $q_i$ a vertical sequence obtained from the $i$-th node. Horizontal sequences $p_t$ and $p_{t+\delta}$ are compared, with $1 \leq \delta \leq T$. In a similar fashion, vertical sequences $q_i$ and $q_{i+\Delta}$ are also compared, with $1 \leq \Delta \leq N$. The similarity obtained from spatial and temporal comparisons are represented by $\mu_H$ and $\mu_V$, respectively. Further experiments regarding the influence the parameters $\Delta$ and $\delta$ are explained in Section S1 of the Supplementary Information.

\subsubsection{LLNA-based pattern recognition}
\label{opadrao}

We employed the LLNA method to extract the intrinsic patterns from textual networks, which aim to distinguish among authors' written style. In the so-called training phase, these techniques first identify patterns for each author's writing style. Then, the patterns identified in the previous phase are used to classify unseen instances in the classification phase. In this manuscript, several well-known supervised classification methods were employed: Bayesian Networks (BNT), Naive Bayes (NVB), RBF Networks (RBF), Multi Layer Perceptron (MLP), Support Vector Machines (SVM), k Nearest Neighbors (kNN), C4.5 (C45) and Random Forest (RFO)~\cite{Bishop:2006:PRM:1162264}. All classifiers were set up with their default configuration of parameters, as suggested in~\cite{10.1371/journal.pone.0094137}.

To evaluate the performance of the classification, we used the k-fold cross-validation strategy~\cite{Bishop:2006:PRM:1162264}. To perform the evaluation, this method splits the data into two sets: the training dataset is the set of samples used for training purposes, while the test set is used for validation purposes. Since these two sets are mutually exclusive and, therefore, the evaluation is performed over unknown instances, the cross-validation method is a reliable strategy. In this study, we use $k=5$ because each author was characterized by a set of 5 books (see description of the dataset in Section~\ref{datasets}). Thus, at each iteration, one book of each author is chosen to compound the test dataset, while the remaining books are selected to form the training dataset.

The results were also further probed by using confusion matrices, which are structures, reporting for each possible class (in our case, for each distinct author) the relationship between predicted and real classes. Traditionally, a confusion matrix is used to identify the following patterns of performance: $\alpha_{m_i,m_i}$, which is the number of instances belonging to class $m_i$ which were correctly assigned to $m_i$; while $\alpha_{m_i,m_j}$ is the number of instances belonging to class $m_i$ which were incorrectly assigned to class $m_j$. Specially, the quantity $\alpha_{m_i,m_j}$ will be useful to identify which authors cannot be discriminated with the proposed technique.

\subsection{Dataset}
\label{datasets}

An English corpus of known authors (labeled instances in the supervised training phase) was created to evaluate the accuracy of the proposed method. The corpus comprises $100$ books, which were extracted from the Project Gutenberg repository~\footnote{See www.gutenberg.org}. The books in our dataset were written by $20$ distinct authors. The full list of books and the respective authors is provided in Section S2 of the Supplementary Information. The distribution of books for authors is uniform, i.e. each author is represented by a set of 5 books. In this study, we considered the task of discriminating among 8 distinct authors. This dataset is hereafter referred to as \textit{validation-dataset}.
Note that datasets using a similar distribution of authors and genres have been considered in related works~\cite{amancio2011comparing,ebrahimpour2013automated,amancio2015concentric,amancio2015authorship}.
The remaining set of $12$ authors, hereafter referred to as \emph{rule-selection-dataset}, was used to the particular process of selecting the best LLNA set of rules.
Note that the choice of best rules was performed in a different dataset because, if the same dataset was used for selecting rules and evaluating classifiers, the obtained results could not represent a true classifier generalization~\cite{Bishop:2006:PRM:1162264}.

In the general scenario of textual classification, the application of pre-processing steps may be useful for the task in hand. In semantical tasks, such as the word sense disambiguation, the lemmatization of words plays an important role on the performance~\cite{Navigli:2009:WSD:1459352.1459355}. In the authorship attribution task, conversely, this same lemmatization step may lead to a great loss of information, hindering the accurate identification of authors' particular writing choices~\cite{ASI:ASI21001}. However, it has been shown that in network based techniques, the lemmatization step is important to cluster distinct writing forms into the same node. In our experiments, we also evaluated three types of lemmatization strategies to generate the textual networks, which led to the creation of three distinct variations of datasets for both \textit{validation-dataset} and \textit{rule-selection-dataset}.
%s
\begin{enumerate}
	\item \textit{none-dataset}: the original dataset was kept, i.e. the lemmatization step was disregarded.
	\item \textit{partial-dataset}: the lemmatization was applied only in nouns. Thus, all nouns are mapped to their singular forms.
	\item \textit{full-dataset}: the lemmatization was applied to all words. Therefore, verbs and nouns are mapped to their infinitive and singular forms, respectively.
\end{enumerate}

\section{Results and discussion}
\label{sec:results}

The main purpose of this manuscript is to characterize networks representing written texts to obtain informative features for the authorship attribution task. Differently from traditional approaches, here we explored the use of LLNA rules to discriminate network topologies. We have used this approach because it has been shown that authors' particular writing choices modify word adjacency networks in a consistent form~\cite{amancio2011comparing}.

As described in Section~\ref{datasets}, our dataset comprises $100$ books written by 20 distinct authors, and three distinct pre-processing strategies were probed to generate the textual networks. In Section \ref{STEvolution}, we qualitatively discuss the patterns arising from the dynamics of the LLNA modelling for each book. In Section \ref{selecRules}, we perform the selection of the best LLNA rules, which are then applied in the authorship problem described in Section \ref{classification}. In Section \ref{other-measures}, we compared the proposed approach with the one based on traditional topological measurements~\cite{amancio2011comparing}. Finally, in Section \ref{sec:EffectLemmatization}, we explore the effects of the lemmatization process on the properties of the networks.

\subsection{LLNA spatio-temporal pattern}
\label{STEvolution}

Table~\ref{tab:spacetime} shows the spatio-temporal diagram of 40 networks of the \textit{partial-dataset} using rule B024678-S4. A spatio-temporal diagram is the representation of the states along time, thus, each column represents the state of a given node and each line represents one time step.
In this particular case, for each spatio-temporal diagram, the columns were ordered by the node degree. Thus, the left-most columns are the nodes taking the lowest degrees $k$, and, the right-most, the ones taking the largest values of $k$. Note that the number of nodes $N$ varies across networks (also reported in Figure~\ref{fig:avg-measures}), therefore, the diagrams are formed by a different number of columns. For simplicity's sake, the diagrams were scaled to fit within the columns of the table.

\begin{table}[h!]
\caption{Spatio-temporal diagrams using the LLNA rule B024678-S4 obtained from books written by eight authors. The \textit{partial-dataset} was used in this case. The LLNA dynamics was performed until $t=500$ and the initial states $s_0$ were defined by a random uniform distribution. The spatio-temporal diagram shows the nodes' states: dead, in black; and alive, in white. While the horizontal axis represent the nodes (sorted by increasing order of degree $k$), the vertical axis represents the temporal variable.}
\label{tab:spacetime}
\centering
\includegraphics[scale=0.71]{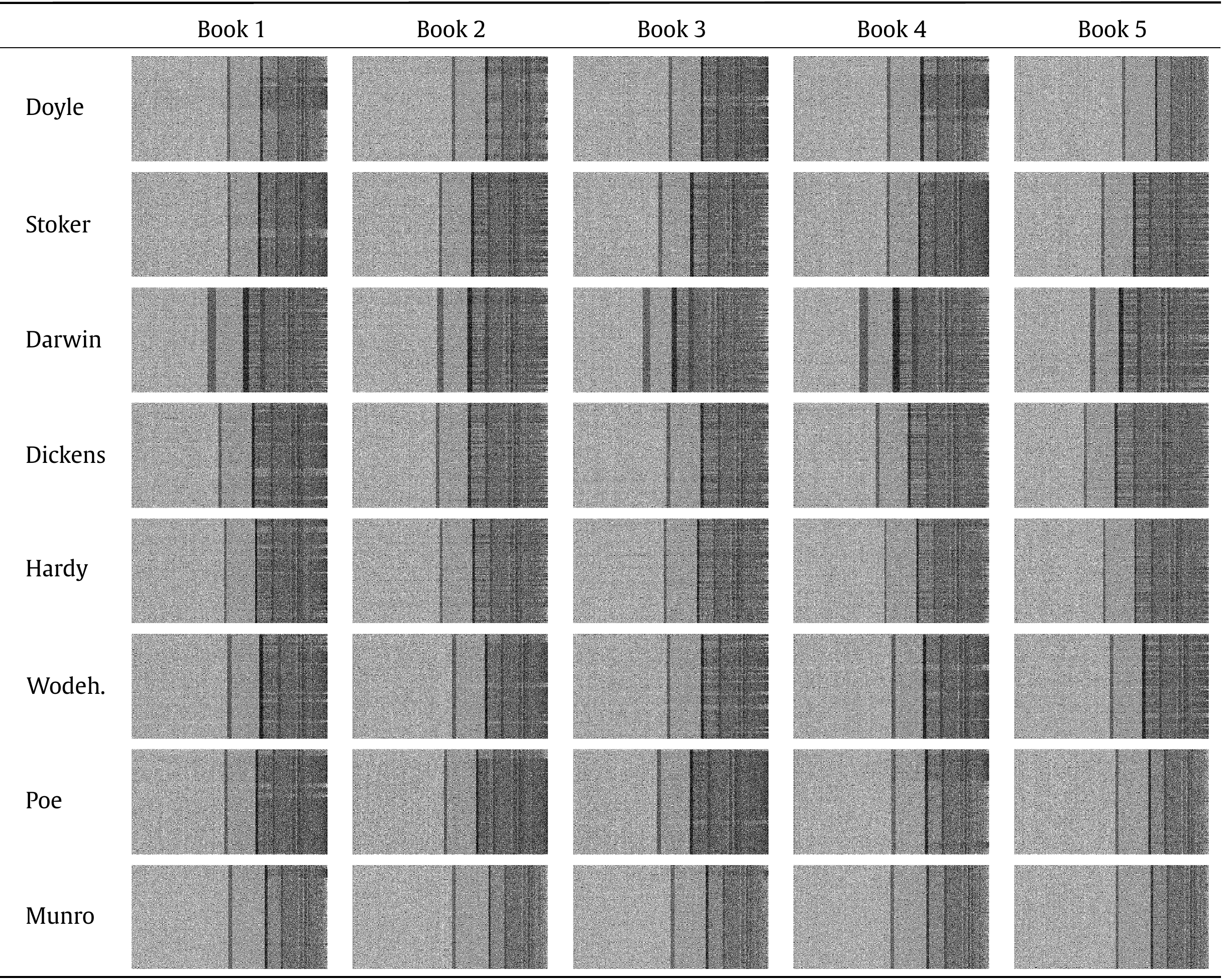}
\end{table}

Notice that for the particular LLNA rule B024678-S4, Table~\ref{tab:spacetime} reveals a general pattern among all the authors. Three notable regions arise: the leftmost correspond to an oscillatory pattern with a higher tendency of alive nodes, followed by a row with tendency of dead nodes (region comprising nodes with average degree $\langle k\rangle =3$). Then, another shorter oscillatory region appears, followed by a second region, which also presents a higher frequency of dead nodes (region comprising nodes with average degree $\langle k\rangle =5$). The reader should note that rule B024678-S4 does not favor nodes with average degrees $3$ and $5$ for birth and survival conditions, which explains the distribution of these vertical patterns in the diagrams. The influence of this rule over the nodes with average degree $1$ is less apparent due to the lower frequency of these nodes. The rightmost nodes, which correspond to hubs in the network, also show oscillatory patterns that are directly related to the dynamics of rule B024678-S4, which favors the birth of the nodes and penalizes their survival. Therefore, there are a dependency between the rule and the network topology.

Despite the above mentioned similar structures in the spatio-temporal diagram, author-dependent patterns can also be noted. For instance, the patterns obtained for Darwin in all five books are strongly similar. Darwin's textual networks present a bigger region corresponding to nodes with average degree $\langle k \rangle=3$, and a major ratio of nodes with high connectivity which are influenced by the rule. Therefore, the spatio-temporal diagram lead us to deduce that the books written by the same author exhibits similar patterns, while allowing to distinguish among the other authors, and that there is a strong dependency of the LLNA's rule.

Based on the spatio-temporal diagram displayed in Table~\ref{tab:spacetime}, we applied measurements (see Section~\ref{Method:LLNAmeasurements}) that allow the characterization of the textual networks in terms of a time series containing only zeros and ones. Before presenting the results of the classification based on time series analysis in Section~\ref{classification}, we first address the LLNA rule selection in the next section.

\subsection{LLNA rule selection}
\label{selecRules}

The rule selection is as important parameter to achieve higher accuracies using the LLNA method~\cite{lifelikeNA}. We evaluated, exhaustively, each of the $262,144$ possible Life-Like rules using the \textit{rule-selection-dataset} comprising 12 authors. As discussed before, the reader should note that the rule selection was performed in different dataset in order to obtain LLNA rules that best represent a true classifier generalization~\cite{Bishop:2006:PRM:1162264}.

To characterize the dynamics of the LLNA, we used a feature vector storing the Shannon entropy and the Lempel-Ziv distributions $[\vec{\mu}_S, \vec{\mu}_L]$, during $t=400$ time steps. Because the choice of the best rule encompasses the induction and evaluation of $262,144$ classifiers, we only used in this phase the kNN method. We have chosen particularly this method because, in general, it generates better results while keeping an excellent processing time~\cite{Bishop:2006:PRM:1162264}. Note that, the application of other methods in this phase, such as neural networks or SVM, would be impractical owing to the time complexity associated to these methods~\cite{Bishop:2006:PRM:1162264}.

\begin{figure}[h]
	\centering
 \includegraphics[scale=0.72]{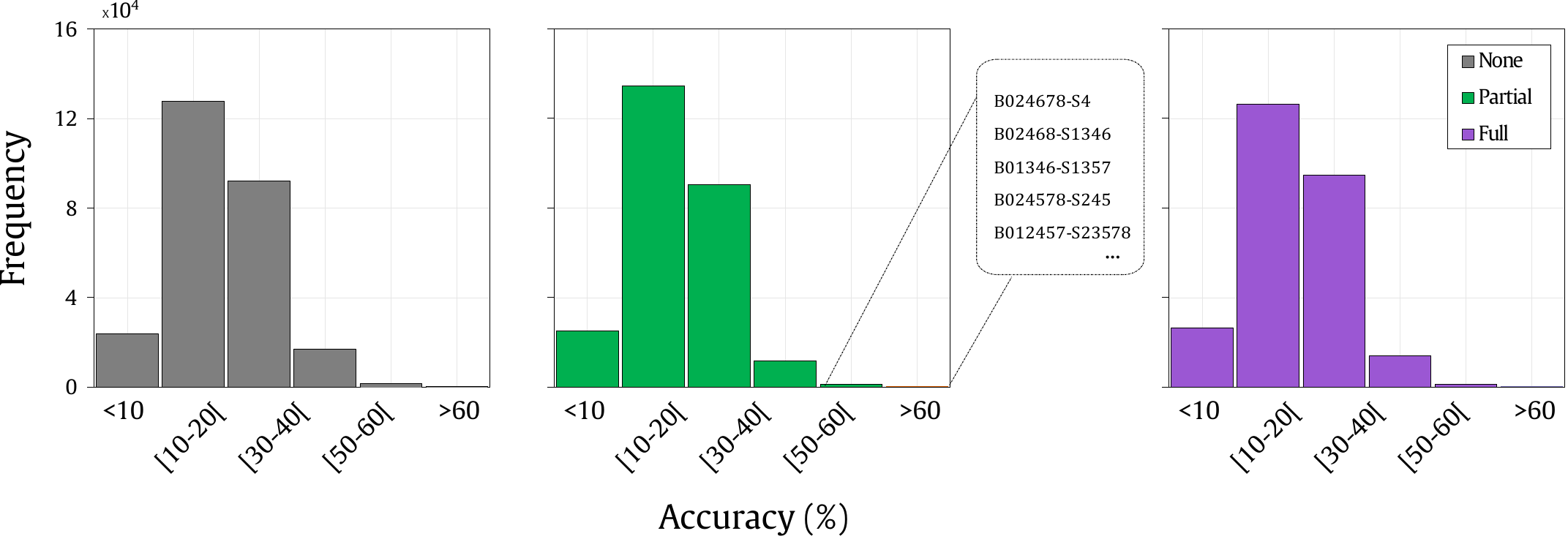}
 \caption{Histogram of the distribution of accuracy for all $262,144$ evaluated rules of the LLNA in the \textit{rule-selection-dataset} comprising 12 authors. From left to right, the histograms for each of the 3 datasets \textit{none}, \textit{partial} and \textit{full}, are shown respectively. As an example, the highlighted five rules maximizes the classification of the \textit{rule-selection-dataset},
 when a partial lemmatization was applied. For this rule selection experiment, both Shannon entropy and Lempel-Ziv complexity were considered as corresponding feature vectors, and kNN classifier.}
	\label{fig:selectionRule}
\end{figure}

Figure~\ref{fig:selectionRule} depicts the histogram distribution of the accuracies obtained for the complete rule-space of the LLNA. Most of the rules yielded low accuracy classifiers. Typically, accuracies lower than 40\% have been found. In this study, we only selected the 400 rules yielding the highest accuracy rates. Note that the selection of best rules is performed independently in each of three datasets: \textit{none-}, \textit{partial-} and \textit{full-} from the \textit{rule-selection-dataset}. Moreover, as the selection rule is a preliminary phase, one should expect that among the set of best rules further improvement can be achieved by using other LLNA measurements~\cite{lifelikeNA}.

\subsection{Classification of authorship networks}
\label{classification}

For the authorship identification problem, we applied the best rules obtained to identify authorship in the \textit{validation-dataset} comprising the 8 authors. First, we compared the three datasets, \textit{none-}, \textit{partial-} and \textit{full-dataset} by using different measurements extracted from the LLNA dynamics: the Shannon entropy distribution $\vec{\mu}_S$, the Lempel-Ziv distribution $\vec{\mu}_L$, and the binary distance distribution, which can be analyzed in a twofold way: horizontally $\vec{\mu}_H$ and vertically $\vec{\mu}_V$ (see Section~\ref{Method:LLNAmeasurements}).

We evaluated the performance of the classification by using different LLNA measurements, extracted from the spatio-temporal pattern, in two ways, isolated and combined. Thus, four feature vectors were used to characterize authors' styles. The first feature vector $\vec{\mu}_S$ is composed by the distribution of the Shannon entropy ${\mu_S}_i$, which is divided into 40 bins, therefore, $\vec{\mu}_S$ contains 40 attributes. Similarly, the second feature vector $\vec{\mu}_L$ is composed by the Lempel-Ziv complexity distribution divided into 40 bins. This vector was normalized by the maximum value achieved among the group of samples. The third and fourth feature vectors are the binary distance distributions, which were explored by means of vertical $\vec{\mu_V}$ and horizontal $\vec{\mu}_H$ analyses, which also contains 30 attributes per measurement. Finally, the combined vector $[\vec{\mu}_S,\vec{\mu}_L,\vec{\mu}_H,\vec{\mu}_V]$ contains 140 attributes.

We tested the accuracy of the 400 selected rules (see Section~\ref{selecRules}) with different feature vectors as well as the combination of them. Table~\ref{tab:acuracia1} presents the best rules obtained for the \textit{validation-dataset}. The columns $\vec{\mu}_S$, $\vec{\mu}_L$, $\vec{\mu}_H$ and $\vec{\mu}_V$ show the accuracy rates obtained for each distinct feature vector. The results when combining these distributions are shown in the last column of the same table. Note that the isolated feature vector $\vec{\mu}_V$ yielded the maximum accuracy of 76.03\% ($\pm$ 12.02\%) for rule B024678-S4 when using the \textit{partial-dataset}.
\setlength{\tabcolsep}{4pt}
\begin{sidewaystable}
\centering
\caption{Accuracy rate (\%) obtained using different measurements and their combinations as attributes to classify 8 authors of the \textit{validation-dataset}. To select the best rules, we used the kNN with k=1 and 5-fold cross validation. The best result among all classifiers (see Section \ref{opadrao}) were also obtained with the kNN method.}
\label{tab:acuracia1}
\begin{tabular}{llccccc}
\toprule
Lemmat. & Rule & $\vec{\mu}_S$ & $\vec{\mu}_L$ & $\vec{\mu}_H$ & $\vec{\mu}_V$ & $[\vec{\mu}_S,\vec{\mu}_L,\vec{\mu}_H,\vec{\mu}_V]$ \\
\midrule
\multirow{3}{*}{\textit{None}} & B03468-S0368 & 45.48 ($\pm$ 11.86) & 39.40 ($\pm$ 13.85) & 22.08 ($\pm$ 13.33) & \textbf{72.75 ($\pm$ 13.54)} & 50.13 ($\pm$ 14.28) \\
 & B138-S3 & 43.95 ($\pm$ 15.82) & 41.23 ($\pm$ 15.02) & 18.58 ($\pm$ 11.32) & 70.88 ($\pm$ 12.94) & 41.73 ($\pm$ 15.32) \\
 & B0124678-S4568 & 40.40 ($\pm$ 14.06) & 47.03 ($\pm$ 14.10) & 43.03 ($\pm$ 16.52) & 68.10 ($\pm$ 14.37) & 53.85 ($\pm$ 16.07) \\
 \midrule
\multirow{3}{*}{\textit{Partial}} & B024678-S4 & 35.58 ($\pm$ 15.74) & 52.63 ($\pm$ 15.74) & 41.18 ($\pm$ 14.65) & \textbf{76.03 ($\pm$ 12.02)} & 49.85 ($\pm$ 16.24) \\
 & B02468-S1346 & 42.85 ($\pm$ 12.24) & 54.35 ($\pm$ 11.54) & 22.75 ($\pm$ 11.79) & 68.25 ($\pm$ 11.28) & 53.75 ($\pm$ 12.09) \\
 & B01346-S1357 & 31.08 ($\pm$ 14.94) & 47.55 ($\pm$ 14.19) & 36.13 ($\pm$ 15.44) & 64.43 ($\pm$ 14.56) & 45.23 ($\pm$ 13.03) \\
 \midrule
\multirow{3}{*}{\textit{Full}} & B1457-S3568 & 36.75 ($\pm$ 10.32) & 46.38 ($\pm$ 14.32) & 9.73 ($\pm$ 8.71) & \textbf{72.72 ($\pm$ 13.20) }& 50.45 ($\pm$ 13.61) \\
 & B15-S278 & 31.68 ($\pm$ 11.61) & 35.85 ($\pm$ 13.81) & 30.30 ($\pm$ 11.77) & 65.80 ($\pm$ 16.13) & 33.83 ($\pm$ 13.31) \\
 & B014568-S13478 & 42.95 ($\pm$ 16.30) & 24.38 ($\pm$ 14.18) & 32.20 ($\pm$ 14.07) & 65.78 ($\pm$ 13.34) & 38.90 ($\pm$ 13.59)\\
\bottomrule
\end{tabular}
\end{sidewaystable}

\begin{figure}[h!]
	\centering
	\includegraphics[scale=0.64]{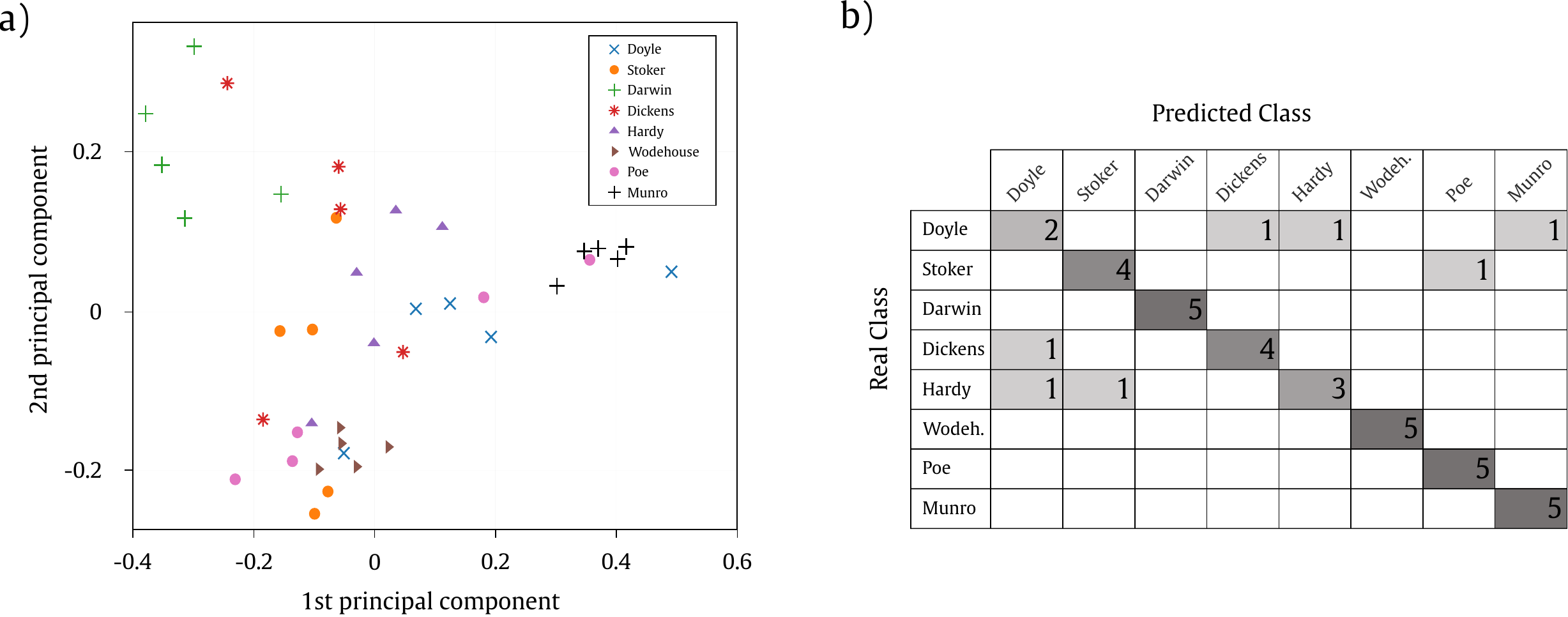}
 \caption{a) Principal component analysis performed for the authorship recognition task using the five books from the authors of the \textit{validation-dataset} using partial lemmatization. For this plot was used rule B024678-S4 and $\vec{\mu}_V$ as a feature vector. b) Confusion matrix using kNN method achieved by the best classification rate. Each cell shows the number of correct predicted instances, where nonzero elements are indicated.}
	\label{fig:pca}
\end{figure}

To illustrate the discriminability obtained with our method, in Figure~\ref{fig:pca}-a), we show a principal component analysis project into two dimensions. In this case, the \textit{partial-dataset} was analyzed, with a dynamics based on the rule B024678-S4 and a characterization performed in terms of the feature vector $\vec{\mu}_V$. Even though only two dimensions were used to visualize our data, there is a clear separation between Darwin and the other authors. A similar pattern occurs for Munro. Interestingly, some authors display a very consistent stylistic (Munro and Wodehouse), while others can considerably vary their styles from book to book (e.g. Dickens).

In Figure \ref{fig:pca}-b) we provide the confusion matrix obtained with the best rule. As expected, Darwin is easily distinguished from the other authors. In a similar fashion, the induced classifier can perfectly discriminate among Wodehouse, Darwin, Poe, and Munro. The author with the lowest classification accuracy is Doyle, since three of his books were incorrectly assigned to Dickens, Hardy and Munro.

The best accuracy rate found using the best configuration of parameters shows unequivocally that the proposed features can capture authors' particularities in written styles, allowing thus the discrimination of authors in unknown texts. Note that, a random authorship attribution would accurately recognize authors with probability $p=1/8=0.125$ in our dataset comprising $n_b=40$ books. Thus, the $p$-value associated with the obtained accuracy of $n_a=33$ books accurately classified (see Figure~\ref{fig:pca}-b)) is
\begin{equation}
	\textrm{$p$-value} = \sum_{n_a=33}^{n_b}  {n_b \choose n_a} p^{n_a} (1-p)^{(n_b-n_a)} \leq 1.0 \times 10^{-15},
\end{equation}
confirming thus the significance of the obtained results. In the next section, we probe the relevance of our results in comparison with the traditional characterization relying only on topological measurements of networks.

\subsection{Evaluation of traditional measurements and robustness analysis}
\label{other-measures}

\begin{figure}[h!]
	\centering
	\includegraphics[scale=0.76]{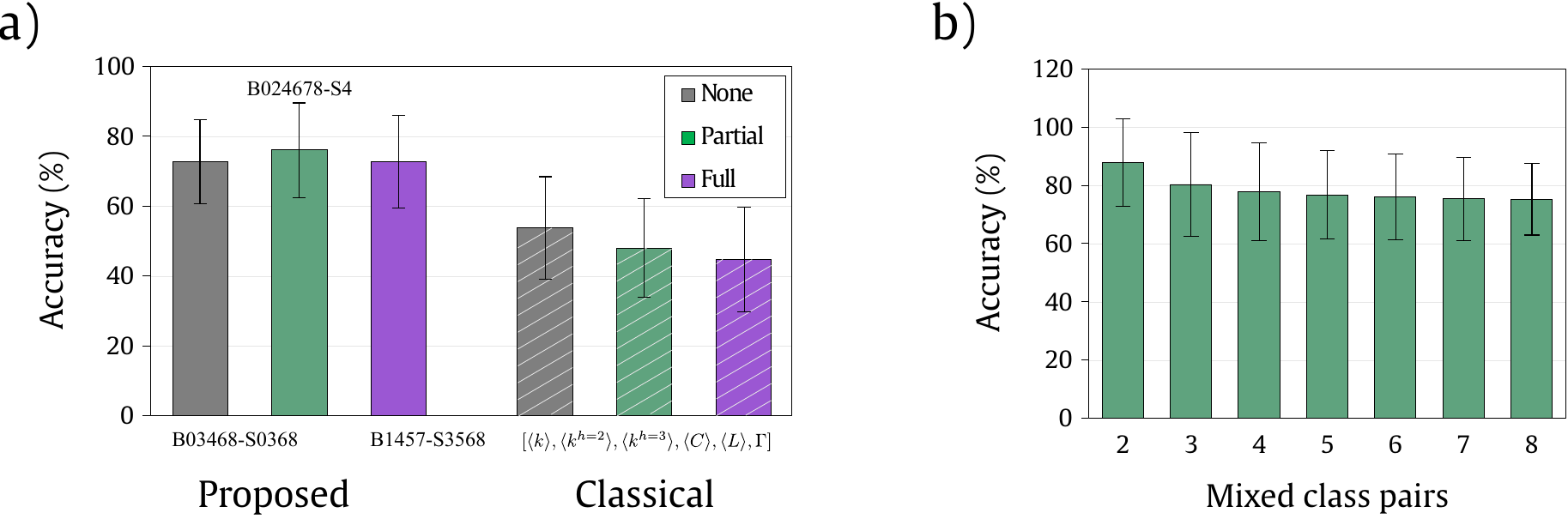}
 \caption{a) Comparison of the accuracy obtained by the proposed method (left side) and the classical network measurements (right side). The histograms on the left (mean and standard deviation) represent the best accuracies obtained when using rules B03468-S0368, B024678-S4 and B1457-S3568 for \textit{none-}, \textit{partial-} and \textit{full-dataset}, respectively. In a similar way, the histograms on the right show the best accuracies obtained using network measurements $[\langle k\rangle, \langle k^{h=2}\rangle, \langle k^{h=3}\rangle, \langle C\rangle, \langle L\rangle,\Gamma]$ as a feature vector. For all these experiments kNN method was used. b) Average accuracy obtained in the variations of the original dataset. Each variation considers a different number of authors, which ranges from $2$ to $8$.}
	\label{fig:comparisonsacuracies}
\end{figure}

We compared the results obtained with the Life-Like network automata with traditional measurements used to characterize complex networks~\cite{amancio2011comparing}. The left side of Table~\ref{table:comparisonaccuracies} shows the accuracy obtained in the classification of the network models when using traditional network measurements. Note that the performance of the traditional method, in general, is improved when no lemmatization is applied. The best result was obtained with the SVM classifier ($61.30\%$), which is similar to the best results reported in~\cite{amancio2011comparing}. A similar performance was also obtained with the MLP classifier ($59.23$\%).
The right side of Table~\ref{table:comparisonaccuracies} shows the results obtained with the proposed method. Rules B03468-S0368, B024678-S4, B1457-S3568 provided the highest accuracies for the \textit{none-}, \textit{partial-} and \textit{full-dataset} when using only the binary distance distribution $\vec{\mu}_V$.
Considering all the variations of both datasets and classifiers, the highest accuracy rate was $76.03\%$. This means that our method outperformed the traditional technique by a margin of 14.73\%, when comparing the best configuration of both strategies. The best results obtained by each strategy are also illustrated in Figure~\ref{fig:comparisonsacuracies}-a).
\begin{sidewaystable}
\centering
\caption{Comparison of the accuracy rate (\%) obtained using traditional network measurements and the proposed method based on network automata. Remarkably, our method outperforms the traditional approach by an average margin of $14.73\%$.}
\label{table:comparisonaccuracies}
\begin{tabular}{llll|lll}
\toprule
 & \multicolumn{3}{c}{Traditional network measurements} & \multicolumn{3}{c}{Proposed method (LLNA)} \\
 \midrule
 & \multicolumn{1}{c}{\textit{None}} & \multicolumn{1}{c}{\textit{Partial}} & \multicolumn{1}{c}{\textit{Full}} & \multicolumn{1}{c}{\begin{tabular}[c]{@{}c@{}}\textit{None}\\ (B03468-S0368)\end{tabular}} & \multicolumn{1}{c}{\begin{tabular}[c]{@{}c@{}}\textit{Partial}\\ (B024678-S4)\end{tabular}} & \multicolumn{1}{c}{\begin{tabular}[c]{@{}c@{}}\textit{Full}\\ (B1457-S3568)\end{tabular}} \\
\midrule
BN & 48.23 ($\pm$14.88) & 45.43 ($\pm$14.18) & 44.23 ($\pm$13.62) & 65.58 ($\pm$14.30) & 44.73 ($\pm$12.86) & 50.55 ($\pm$14.30) \\
NVB & 58.28 ($\pm$15.16) & 56.23 ($\pm$14.5) & 51.13 ($\pm$15.07) & 62.80 ($\pm$15.24) & 57.48 ($\pm$15.37) & 50.10 ($\pm$14.42) \\
MLP & 59.23 ($\pm$13.92) & 50.73 ($\pm$14.13) & 45.03 ($\pm$15.2) & 69.63 ($\pm$14.36) & 59.25 ($\pm$14.11) & 60.50 ($\pm$14.41) \\
KNN & 52.00 ($\pm$14.9) & 49.40 ($\pm$14.52) & 43.65 ($\pm$16.38) & 72.75 ($\pm$13.54) & \textbf{76.03 ($\pm$12.02)} & 72.72 ($\pm$13.20) \\
C45 & 44.25 ($\pm$13.05) & 42.55 ($\pm$14.72) & 42.13 ($\pm$14.47) & 52.15 ($\pm$14.56) & 32.53 ($\pm$13.29) & 45.05 ($\pm$15.22) \\
RF & 53.43 ($\pm$15.01) & 54.60 ($\pm$14.44) & 45.88 ($\pm$14.18) & 69.45 ($\pm$12.70) & 61.25 ($\pm$14.5) & 63.32 ($\pm$14.51) \\
RBF & 52.48 ($\pm$14.17) & 51.68 ($\pm$14.68) & 46.38 ($\pm$15.52) & 27.30 ($\pm$6.83) & 51.15 ($\pm$15.67) & 38.40 ($\pm$8.33) \\
SVM & \textbf{61.30 ($\pm$15.56)} & 49.28 ($\pm$13.97) & 50.20 ($\pm$14.70) & 72.65 ($\pm$12.69) & 70.03 ($\pm$13.38) & 66.45 ($\pm$14.13)\\
\bottomrule
\end{tabular}
\end{sidewaystable}

The robustness of the proposed methodology with regard to the total number of authors considered was verified by considering other variations of authors in the \textit{validation-dataset}. To do so, we selected all variations of $8$ authors among the total of $20$ authors. We then applied the proposed methodology to probe the sensibility of our method to specific datasets. As shown in Figure \ref{fig:comparisonsacuracies}-b), there is only a minor variation in the accuracy when considering datasets of $8$ authors, suggesting that our method is robust with regard to the variation of datasets. A similar procedure was performed to study the robustness in datasets comprising a distinct number of authors (from 2 to 7 authors). Note that, in these other scenarios, a similar robust behavior was found. Interestingly, similar accuracy results have been obtained when considering 3 and 8 authors, suggesting thus that our method is more effective when more complex authorship attribution tasks are considered.

\subsection{Effect of the lemmatization on network measurements}
\label{sec:EffectLemmatization}

Table~\ref{Tab:metricsAuthor} shows the preliminary topological properties for one of Doyle's book modeled as a network, considering the three lemmatization processes (\textit{none}, \textit{partial} and \textit{full}). The columns show the measurements presented in Section~\ref{Method:networkmeasurements}, as follows: number of nodes $N$, number of edges $E$, average degree $\langle k\rangle$, clustering coefficient $\langle C\rangle$, average path length $\langle L\rangle$, power-law exponent $\gamma$, diameter $D$, density $d$ and degree assortativity $\Gamma$.

From the same table, one can note a decreasing of both the number of nodes $N$ and edges $E$, while the average degree $\langle k\rangle$ increases. This effect can be explained by the fact that when the lemmatization process is performed, the multiple representations of a word are all transformed to its canonical form, e.g., the words \textit{has} and \textit{have} will have only one representation in a network, the node \textit{have}, instead of having two. Moreover, the diameter for all the networks is maintained around 11. We also observed that all networks studied here obey a power law constant around $\approx 2.37$. Therefore, these textual networks have a scale-free structure, which is supported by the maximum likelihood method and the Kolmogorov-Smirnov statistic that accepts the hypotheses of a reasonable fit. Moreover, this property is consistent with the scale-free textual networks found in the literature.
\begin{table}[h!]
\centering
\caption{Measurements extracted for the textual network corresponding to Doyle's book ``\textit{Uncle Bernac - A Memory of the Empire}" regarding the three types of lemmatization process (\textit{none-}, \textit{partial-} and \textit{full-dataset}).}
\label{Tab:metricsAuthor}
\begin{tabular}{lllllllllll}
\toprule
Lemm. & \multicolumn{1}{c}{$N$} & \multicolumn{1}{c}{$E$} & \multicolumn{1}{c}{$\langle k\rangle$} & \multicolumn{1}{c}{$\langle C\rangle$} & \multicolumn{1}{c}{$\langle L \rangle$} & \multicolumn{1}{c}{$\gamma$} & \multicolumn{1}{c}{$D$} & \multicolumn{1}{c}{$d$} & \multicolumn{1}{c}{$\Gamma$} \\
\midrule
\textit{None} & 5914 & 22991&7.78&0.04&3.63&2.33&11&0.0013&-0.06\\
\textit{Partial}&5374&22775&8.48&0.04&3.54&2,29&11&0.0016&-0.06\\
\textit{Full}&4977&22451&9.02&0.05&3.47&2.20&10&0.0018&-0.07\\
\bottomrule
\end{tabular}
\end{table}
\begin{figure}[h!]
	\centering
	\includegraphics[scale=0.6]{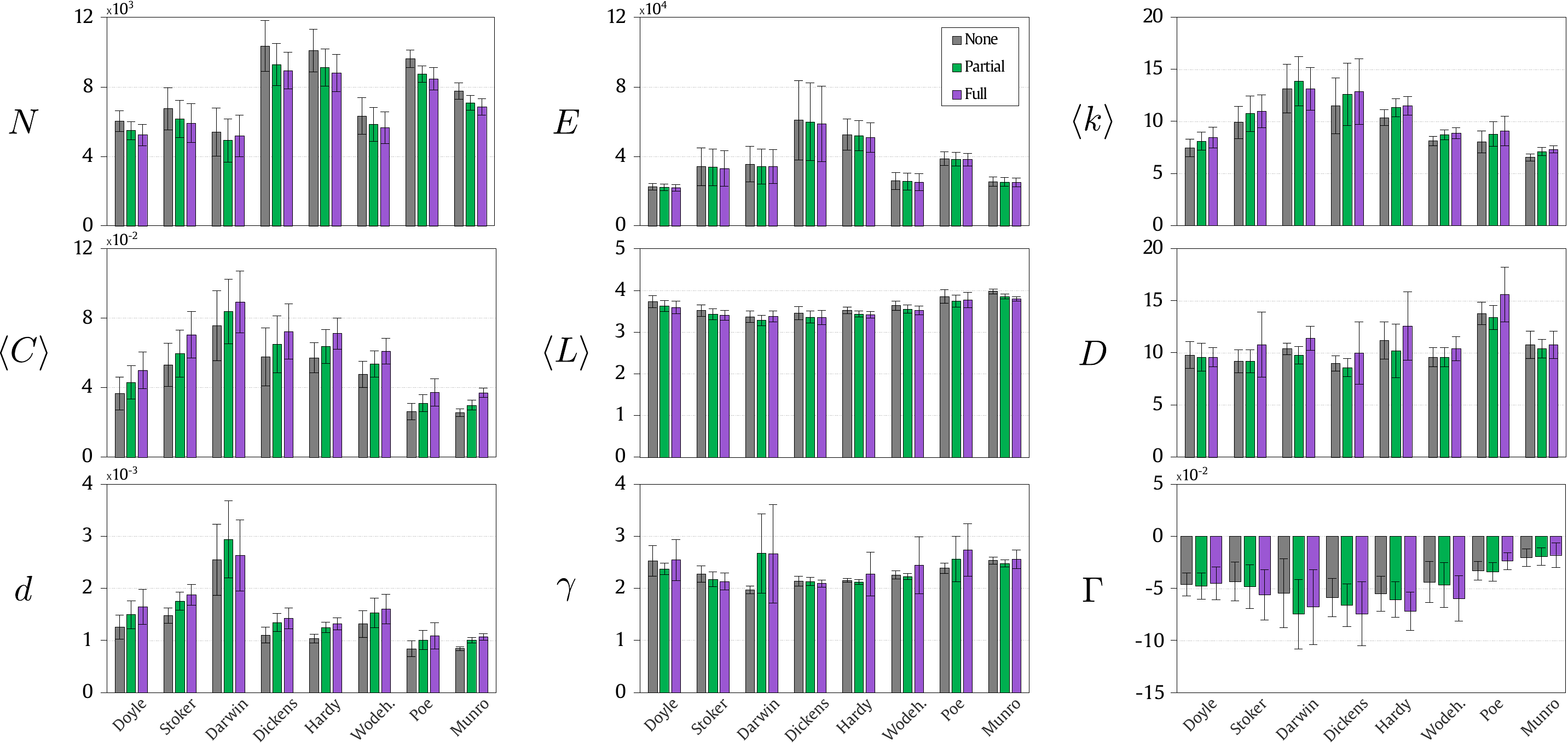}
 \caption{Average network measurements for eight authors highlighted in the diagrams and for the three datasets: \textit{none-}, \textit{partial-} and \textit{full-dataset} (see description in Section \ref{datasets}). The following distributions are shown for each author: number of nodes ($N$), number of edges ($E$), average connectivity ($\langle k\rangle$), average clustering coefficient ($\langle C \rangle$), average path length ($\langle L\rangle$), diameter ($D$), density ($d$), power-law exponent ($\gamma)$ and degree assortativity ($\Gamma$).}
	\label{fig:avg-measures}
\end{figure}

Figure~\ref{fig:avg-measures} presents a set of average topological measurements calculated for each author of the \textit{validation-dataset}. The standard deviation was obtained considering the five books of each author. Figure~\ref{fig:avg-measures} also shows the values obtained for the three variations of dataset. The main results concerning each measurement are described below:

\begin{itemize}
\item {\bf Total number of nodes ($N$) and edges ($E$)}: $N$ decreases with the lemmatization process, whereas $E$ is not influenced by this process. This effect occurs because, even when nodes are removed during the lemmatization, adjacency relationships are not affected, and, consequently, the degree of the remaining nodes tends to increase. This effect is evident in the top-right diagram displaying the average network connectivity $\langle k \rangle$.

\item {\bf Average clustering coefficient ($\langle C \rangle$)}: This measurement was influenced by both $N$ and $k$. $\langle C \rangle$ tends to increase with the lemmatization process because the network remains with almost the same number of edges, while the number of nodes decreases as a consequence of mapping distinct variations of the same concept into the same node.
\item {\bf Average shortest path length ($\langle L\rangle$)}: Similarly to the number of edges, the average shortest path length is not much affected by the lemmatization process. However, note that the values of $\langle L\rangle$ tend to decrease as a consequence of the decrease in the total number of nodes.
\item {\bf Diameter ($D$)}: In most cases, the diameter increases by a short margin when the lemmatization process is performed. However, this pattern seems to depend from author to author. Note, e.g. that the average diameter decreases when the full lemmatization is applied for books authored by Doyle. Conversely, the lemmatization process seems to cause an opposite effect on networks modelling books written by Allan Poe.
\item {\bf Density ($d$)}: The density of links increases in most cases, as the lemmatization process removes nodes, and the number of edges is practically not affected. An exception occurs for Darwin. Remarkably, the average density of the \textit{none-} and \textit{full-} datasets are in a similar fashion.
\item {\bf Power-law exponent ($\gamma)$}: Almost all the textual networks present power exponent between 2 and 3, which is a characteristic that have been demonstrated for many real-world networks~\cite{Newman:2010:NI:1809753,dorogovtsev2002evolution} and, particularly in text networks, is a consequence of the Zipf's Law. Concerning the effect of the lemmatization process on this feature, no clear pattern can be identified, as opposite effects have been found e.g. for Stoker and Poe.
\end{itemize}

\section{Conclusion}
\label{sec:conclusion}

In this paper, we have addressed the authorship attribution problem, which is a task of practical relevance in many contexts of information science research. We have specifically studied the effect of the textual organization in the discriminability of documents written by distinct authors. To capture the structural properties of texts, we have used the well-known network framework, given its potential revealed in related applications. Unlike the traditional approach based only on topological properties of networks, we have proposed here a methodology to capture further information concerning authors' particular styles. To do so, we have represented networks modelling texts as network automata with a dynamics based on Life-Like rules. Upon selecting a set of discriminative rules that serve to coordinate the automata dynamics, we have found that the variations in the binary states of nodes are more discriminative than simple traditional topological characterization. More specifically, we have obtained an improvement of almost 15\% in the classification of $8$ distinct authors. Interestingly, the best results were obtained with a partial lemmatization process, suggesting that this procedure is more adequate than just lemmatizing all words when text networks are used as the underlying model for this task.

The methodology proposed here paves the way for improving the characterization of related information systems modelled in terms of networks. This is evident if we recall that network automata approaches are specially suitable to describe networks with scale-free distributions~\cite{lifelikeNA} and, as a consequence, documents following Zipf's Law. Further works could investigate the effectiveness of our approach e.g. in the analysis of the complexity of texts or in applications related to extractive summarization. Given the complementarity of the analysis provided by the network automata framework, we argue that a combination relying on traditional superficial and networked features could lead to optimized results in a variety of natural language processing applications.

\section*{Acknowledgements}
%Acknowledgements should be brief, and should not include thanks to anonymous referees and editors, or effusive comments. Grant or contribution numbers may be acknowledged.

% grant #aaaa/nnnnn-d, São Paulo Research Foundation (FAPESP).
J.M. is grateful for the support of the Coordination for the Improvement of Higher Education Personnel (CAPES).
E.A.C.J. and D.R.A. are grateful for the support from Google (Google Research Awards in Latin America grant).
D.R.A. is also grateful for the financial support from S\~{a}o Paulo Research Foundation (FAPESP grant \#2014/20830-0).
G.H.B.M. is grateful for the support from CAPES and FAPESP with grant \#2015/05899-7.
O.M.B. gratefully acknowledges the financial support of CNPq (National Council for Scientific and Technological Development, Brazil) (Grant \#307797/2014-7 and Grant \#484312/2013-8) and FAPESP (Grant \#11/01523-1 and Grant \#2015/05899-7).

%%%% VVVVVVV This is the correct refs. format for Information Sciences VVVVVVVV
%\bibliography{references}

\begin{thebibliography}{10}
\expandafter\ifx\csname url\endcsname\relax
  \def\url#1{\texttt{#1}}\fi
\expandafter\ifx\csname urlprefix\endcsname\relax\def\urlprefix{URL }\fi
\providecommand{\bibinfo}[2]{#2}
\providecommand{\eprint}[2][]{\url{#2}}

\bibitem{FrancoSalvador2016550}
\bibinfo{author}{Franco-Salvador, M.}, \bibinfo{author}{Rosso, P.} \&
  \bibinfo{author}{{Montes-y-G\'{o}mez}, M.}
\newblock \bibinfo{title}{A systematic study of knowledge graph analysis for
  cross-language plagiarism detection}.
\newblock \emph{\bibinfo{journal}{Information Processing \& Management}}
  \textbf{\bibinfo{volume}{52}}, \bibinfo{pages}{550--570}
  (\bibinfo{year}{2016}).

\bibitem{Labbe}
\bibinfo{author}{Labb{\'e}, C.} \& \bibinfo{author}{Labb{\'e}, D.}
\newblock \bibinfo{title}{Duplicate and fake publications in the scientific
  literature: how many scigen papers in computer science?}
\newblock \emph{\bibinfo{journal}{Scientometrics}}
  \textbf{\bibinfo{volume}{94}}, \bibinfo{pages}{379--396}
  (\bibinfo{year}{2013}).

\bibitem{Vacca:2005:CFC:1076307}
\bibinfo{author}{Vacca, J.~R.}
\newblock \emph{\bibinfo{title}{Computer Forensics: Computer Crime Scene
  Investigation (Networking Series) (Networking Series)}}
  (\bibinfo{publisher}{Charles River Media, Inc.}, \bibinfo{address}{Rockland,
  MA, USA}, \bibinfo{year}{2005}).

\bibitem{ASI:ASI21001}
\bibinfo{author}{Stamatatos, E.}
\newblock \bibinfo{title}{A survey of modern authorship attribution methods}.
\newblock \emph{\bibinfo{journal}{Journal of the American Society for
  Information Science and Technology}} \textbf{\bibinfo{volume}{60}},
  \bibinfo{pages}{538--556} (\bibinfo{year}{2009}).

\bibitem{amancio2015authorship}
\bibinfo{author}{Amancio, D.~R.}
\newblock \bibinfo{title}{Authorship recognition via fluctuation analysis of
  network topology and word intermittency}.
\newblock \emph{\bibinfo{journal}{Journal of Statistical Mechanics: Theory and
  Experiment}} \textbf{\bibinfo{volume}{2015}}, \bibinfo{pages}{P03005}
  (\bibinfo{year}{2015}).

\bibitem{Brennan:2012:ASC:2382448.2382450}
\bibinfo{author}{Brennan, M.}, \bibinfo{author}{Afroz, S.} \&
  \bibinfo{author}{Greenstadt, R.}
\newblock \bibinfo{title}{Adversarial stylometry: Circumventing authorship
  recognition to preserve privacy and anonymity}.
\newblock \emph{\bibinfo{journal}{ACM Trans. Inf. Syst. Secur.}}
  \textbf{\bibinfo{volume}{15}}, \bibinfo{pages}{12:1--12:22}
  (\bibinfo{year}{2012}).

\bibitem{Halteren:2007:AVL:1187415.1187416}
\bibinfo{author}{Halteren, H.~V.}
\newblock \bibinfo{title}{Author verification by linguistic profiling: an
  exploration of the parameter space}.
\newblock \emph{\bibinfo{journal}{ACM Trans. Speech Lang. Process.}}
  \textbf{\bibinfo{volume}{4}}, \bibinfo{pages}{1--17} (\bibinfo{year}{2007}).

\bibitem{Martinici}
\bibinfo{author}{Martincic-Ipsic, S.}, \bibinfo{author}{Margan, D.} \&
  \bibinfo{author}{Mestrovic, A.}
\newblock \bibinfo{title}{Multilayer network of language: a unified framework
  for structural analysis of linguistic subsystems}.
\newblock \emph{\bibinfo{journal}{Physica A: Statistical Mechanics and its
  Applications}} \textbf{\bibinfo{volume}{457}}, \bibinfo{pages}{117--128}
  (\bibinfo{year}{2016}).

\bibitem{Dorogovtsev2603}
\bibinfo{author}{Dorogovtsev, S.~N.} \& \bibinfo{author}{Mendes, J. F.~F.}
\newblock \bibinfo{title}{Language as an evolving word web}.
\newblock \emph{\bibinfo{journal}{Proceedings of the Royal Society of London B:
  Biological Sciences}} \textbf{\bibinfo{volume}{268}},
  \bibinfo{pages}{2603--2606} (\bibinfo{year}{2001}).

\bibitem{0295-5075-100-5-58002}
\bibinfo{author}{Amancio, D.~R.}, \bibinfo{author}{Aluisio, S.~M.},
  \bibinfo{author}{Oliveira~Jr., O.~N.} \& \bibinfo{author}{Costa, L.~F.}
\newblock \bibinfo{title}{Complex networks analysis of language complexity}.
\newblock \emph{\bibinfo{journal}{EPL (Europhysics Letters)}}
  \textbf{\bibinfo{volume}{100}}, \bibinfo{pages}{58002}
  (\bibinfo{year}{2012}).

\bibitem{10.1371/journal.pone.0067310}
\bibinfo{author}{Amancio, D.~R.}, \bibinfo{author}{Altmann, E.~G.},
  \bibinfo{author}{Rybski, D.}, \bibinfo{author}{Oliveira~Jr., O.~N.} \&
  \bibinfo{author}{Costa, L.~F.}
\newblock \bibinfo{title}{Probing the statistical properties of unknown texts:
  application to the voynich manuscript}.
\newblock \emph{\bibinfo{journal}{PLoS ONE}} \textbf{\bibinfo{volume}{8}},
  \bibinfo{pages}{e67310} (\bibinfo{year}{2013}).

\bibitem{0295-5075-93-2-28005}
\bibinfo{author}{Liu, H.} \& \bibinfo{author}{Xu, C.}
\newblock \bibinfo{title}{Can syntactic networks indicate morphological
  complexity of a language?}
\newblock \emph{\bibinfo{journal}{EPL (Europhysics Letters)}}
  \textbf{\bibinfo{volume}{93}}, \bibinfo{pages}{28005} (\bibinfo{year}{2011}).

\bibitem{0295-5075-83-1-18002}
\bibinfo{author}{Liu, H.} \& \bibinfo{author}{Hu, F.}
\newblock \bibinfo{title}{What role does syntax play in a language network?}
\newblock \emph{\bibinfo{journal}{EPL (Europhysics Letters)}}
  \textbf{\bibinfo{volume}{83}}, \bibinfo{pages}{18002} (\bibinfo{year}{2008}).

\bibitem{Mehri20122429}
\bibinfo{author}{Mehri, A.}, \bibinfo{author}{Darooneh, A.~H.} \&
  \bibinfo{author}{Shariati, A.}
\newblock \bibinfo{title}{The complex networks approach for authorship
  attribution of books}.
\newblock \emph{\bibinfo{journal}{Physica A: Statistical Mechanics and its
  Applications}} \textbf{\bibinfo{volume}{391}}, \bibinfo{pages}{2429 -- 2437}
  (\bibinfo{year}{2012}).

\bibitem{amancio2011comparing}
\bibinfo{author}{Amancio, D.~R.}, \bibinfo{author}{Altmann, E.~G.},
  \bibinfo{author}{Oliveira~Jr., O.~N.} \& \bibinfo{author}{Costa, L.~F.}
\newblock \bibinfo{title}{Comparing intermittency and network measurements of
  words and their dependence on authorship}.
\newblock \emph{\bibinfo{journal}{New Journal of Physics}}
  \textbf{\bibinfo{volume}{13}}, \bibinfo{pages}{123024}
  (\bibinfo{year}{2011}).

\bibitem{WOLFRAM19841}
\bibinfo{author}{Wolfram, S.}
\newblock \bibinfo{title}{Universality and complexity in cellular automata}.
\newblock \emph{\bibinfo{journal}{Physica D: Nonlinear Phenomena}}
  \textbf{\bibinfo{volume}{10}}, \bibinfo{pages}{1--35} (\bibinfo{year}{1984}).

\bibitem{watts1999small}
\bibinfo{author}{Watts, D.~J.}
\newblock \emph{\bibinfo{title}{Small worlds: the dynamics of networks between
  order and randomness}} (\bibinfo{publisher}{Princeton university press},
  \bibinfo{year}{1999}).

\bibitem{tomassini2005evolution}
\bibinfo{author}{Tomassini, M.}, \bibinfo{author}{Giacobini, M.} \&
  \bibinfo{author}{Darabos, C.}
\newblock \bibinfo{title}{Evolution and dynamics of small-world cellular
  automata}.
\newblock \emph{\bibinfo{journal}{Complex Systems}}
  \textbf{\bibinfo{volume}{15}}, \bibinfo{pages}{261--284}
  (\bibinfo{year}{2005}).

\bibitem{marr2012cellular}
\bibinfo{author}{Marr, C.} \& \bibinfo{author}{H{\"u}tt, M.-T.}
\newblock \bibinfo{title}{Cellular automata on graphs: Topological properties
  of er graphs evolved towards low-entropy dynamics}.
\newblock \emph{\bibinfo{journal}{Entropy}} \textbf{\bibinfo{volume}{14}},
  \bibinfo{pages}{993--1010} (\bibinfo{year}{2012}).

\bibitem{lifelikeNA}
\bibinfo{author}{Miranda, G.}, \bibinfo{author}{Machicao, J.} \&
  \bibinfo{author}{Bruno, O.~M.}
\newblock \bibinfo{title}{Exploring spatio-temporal patterns as network
  descriptors based on cellular automata}.
\newblock \emph{\bibinfo{journal}{Scientific Reports (under review)}}
  (\bibinfo{year}{2016}).

\bibitem{gonccalves2012complex}
\bibinfo{author}{Gon{\c{c}}alves, W.~N.}, \bibinfo{author}{Martinez, A.~S.} \&
  \bibinfo{author}{Bruno, O.~M.}
\newblock \bibinfo{title}{Complex network classification using partially
  self-avoiding deterministic walks}.
\newblock \emph{\bibinfo{journal}{Chaos: An Interdisciplinary Journal of
  Nonlinear Science}} \textbf{\bibinfo{volume}{22}}, \bibinfo{pages}{033139}
  (\bibinfo{year}{2012}).

\bibitem{GardnerMathematicalGT}
\bibinfo{author}{Gardner, M.}
\newblock \bibinfo{title}{Mathematical games the fantastic combinations of john
  conway's new solitaire game "life"}.
\newblock \emph{\bibinfo{journal}{Scientific American}}
  \textbf{\bibinfo{volume}{223}}, \bibinfo{pages}{120--123}
  (\bibinfo{year}{1970}).

\bibitem{Soto2015TheXU}
\bibinfo{author}{Soto, J. M.~G.} \& \bibinfo{author}{Wuensche, A.}
\newblock \bibinfo{title}{The x-rule: Universal computation in a non-isotropic
  life-like cellular automaton}.
\newblock \emph{\bibinfo{journal}{J. Cellular Automata}}
  \textbf{\bibinfo{volume}{10}}, \bibinfo{pages}{261--294}
  (\bibinfo{year}{2015}).

\bibitem{machicao2012chaotic}
\bibinfo{author}{Machicao, J.}, \bibinfo{author}{Marco, A.~G.} \&
  \bibinfo{author}{Bruno, O.~M.}
\newblock \bibinfo{title}{Chaotic encryption method based on life-like cellular
  automata}.
\newblock \emph{\bibinfo{journal}{Expert Systems with Applications}}
  \textbf{\bibinfo{volume}{39}}, \bibinfo{pages}{12626--12635}
  (\bibinfo{year}{2012}).

\bibitem{Broderick2004ALV}
\bibinfo{author}{Broderick, G.}, \bibinfo{author}{R{\'u}aini, M.},
  \bibinfo{author}{Chan, E.} \& \bibinfo{author}{Ellison, M.~J.}
\newblock \bibinfo{title}{A life-like virtual cell membrane using discrete
  automata}.
\newblock \emph{\bibinfo{journal}{In Silico Biology}}
  \textbf{\bibinfo{volume}{5}}, \bibinfo{pages}{163--178}
  (\bibinfo{year}{2004}).

\bibitem{CsuhajVarj1997EcoGrammarSA}
\bibinfo{author}{Csuhaj-Varj{\'u}, E.}, \bibinfo{author}{Kelemen, J.},
  \bibinfo{author}{Kelemenov{\'a}, A.} \& \bibinfo{author}{Paun, G.}
\newblock \bibinfo{title}{Eco-grammar systems: A grammatical framework for
  studying life-like interaction}.
\newblock \emph{\bibinfo{journal}{Artificial Life}}
  \textbf{\bibinfo{volume}{3}}, \bibinfo{pages}{1--28} (\bibinfo{year}{1997}).

\bibitem{mihalcea2011graph}
\bibinfo{author}{Mihalcea, R.} \& \bibinfo{author}{Radev, D.}
\newblock \emph{\bibinfo{title}{Graph-based natural language processing and
  information retrieval}} (\bibinfo{publisher}{Cambridge University Press},
  \bibinfo{year}{2011}).

\bibitem{sole2010language}
\bibinfo{author}{Sol{\'e}, R.~V.}, \bibinfo{author}{Corominas-Murtra, B.},
  \bibinfo{author}{Valverde, S.} \& \bibinfo{author}{Steels, L.}
\newblock \bibinfo{title}{Language networks: Their structure, function, and
  evolution}.
\newblock \emph{\bibinfo{journal}{Complexity}} \textbf{\bibinfo{volume}{15}},
  \bibinfo{pages}{20--26} (\bibinfo{year}{2010}).

\bibitem{amancio2011using}
\bibinfo{author}{Amancio, D.~R.} \emph{et~al.}
\newblock \bibinfo{title}{Using metrics from complex networks to evaluate
  machine translation}.
\newblock \emph{\bibinfo{journal}{Physica A: Statistical Mechanics and its
  Applications}} \textbf{\bibinfo{volume}{390}}, \bibinfo{pages}{131--142}
  (\bibinfo{year}{2011}).

\bibitem{collins2002discriminative}
\bibinfo{author}{Collins, M.}
\newblock \bibinfo{title}{Discriminative training methods for hidden markov
  models: Theory and experiments with perceptron algorithms}.
\newblock In \emph{\bibinfo{booktitle}{Proceedings of the ACL-02 conference on
  Empirical methods in natural language processing-Volume 10}},
  \bibinfo{pages}{1--8} (\bibinfo{organization}{Association for Computational
  Linguistics}, \bibinfo{year}{2002}).

\bibitem{toman2006influence}
\bibinfo{author}{Toman, M.}, \bibinfo{author}{Tesar, R.} \&
  \bibinfo{author}{Jezek, K.}
\newblock \bibinfo{title}{Influence of word normalization on text
  classification}.
\newblock \emph{\bibinfo{journal}{Proceedings of InSciT}}
  \textbf{\bibinfo{volume}{4}}, \bibinfo{pages}{354--358}
  (\bibinfo{year}{2006}).

\bibitem{Newman:2010:NI:1809753}
\bibinfo{author}{Newman, M. E.~J.}
\newblock \emph{\bibinfo{title}{Networks: An Introduction}}
  (\bibinfo{publisher}{Oxford University Press, Inc.}, \bibinfo{address}{New
  York, NY, USA}, \bibinfo{year}{2010}).

\bibitem{Clauset-PowerLawMatlab}
\bibinfo{author}{Clauset, A.}, \bibinfo{author}{Shalizi, C.~R.} \&
  \bibinfo{author}{Newman, M. E.~J.}
\newblock \bibinfo{title}{Power-law distributions in empirical data}.
\newblock \emph{\bibinfo{journal}{SIAM Rev.}} \textbf{\bibinfo{volume}{51}},
  \bibinfo{pages}{661--703} (\bibinfo{year}{2009}).

\bibitem{Li2016649}
\bibinfo{author}{Li, T.} \emph{et~al.}
\newblock \bibinfo{title}{An epidemic spreading model on adaptive scale-free
  networks with feedback mechanism}.
\newblock \emph{\bibinfo{journal}{Physica A: Statistical Mechanics and its
  Applications}} \textbf{\bibinfo{volume}{450}}, \bibinfo{pages}{649--656}
  (\bibinfo{year}{2016}).

\bibitem{10.1371/journal.pone.0110121}
\bibinfo{author}{Williams, O.} \& \bibinfo{author}{Del~Genio, C.~I.}
\newblock \bibinfo{title}{Degree correlations in directed scale-free networks}.
\newblock \emph{\bibinfo{journal}{PLoS ONE}} \textbf{\bibinfo{volume}{9}},
  \bibinfo{pages}{1--6} (\bibinfo{year}{2014}).

\bibitem{sscoor}
\bibinfo{author}{Morita, S.}
\newblock \bibinfo{title}{Six susceptible-infected-susceptible models on
  scale-free networks}.
\newblock \emph{\bibinfo{journal}{Scientific Reports}}
  \textbf{\bibinfo{volume}{6}}, \bibinfo{pages}{22506 EP}
  (\bibinfo{year}{2016}).

\bibitem{0295-5075-99-2-28002}
\bibinfo{author}{Carron, P.~M.} \& \bibinfo{author}{Kenna, R.}
\newblock \bibinfo{title}{Universal properties of mythological networks}.
\newblock \emph{\bibinfo{journal}{EPL (Europhysics Letters)}}
  \textbf{\bibinfo{volume}{99}}, \bibinfo{pages}{28002} (\bibinfo{year}{2012}).

\bibitem{lantiq}
\bibinfo{author}{Costa, L.~F.}, \bibinfo{author}{Sporns, O.},
  \bibinfo{author}{Antiqueira, L.}, \bibinfo{author}{Nunes, M. G.~V.} \&
  \bibinfo{author}{Oliveira~Jr., O.~N.}
\newblock \bibinfo{title}{Correlations between structure and random walk
  dynamics in directed complex networks}.
\newblock \emph{\bibinfo{journal}{Applied Physics Letters}}
  \textbf{\bibinfo{volume}{91}} (\bibinfo{year}{2007}).

\bibitem{newman2002}
\bibinfo{author}{Newman, M. E.~J.}
\newblock \bibinfo{title}{Assortative mixing in networks}.
\newblock \emph{\bibinfo{journal}{Phys. Rev. Lett.}}
  \textbf{\bibinfo{volume}{89}}, \bibinfo{pages}{208701}
  (\bibinfo{year}{2002}).

\bibitem{shannon19481mathematical}
\bibinfo{author}{Shannon, C.}
\newblock \bibinfo{title}{A mathematical theory of communication}.
\newblock \emph{\bibinfo{journal}{The Bell System Technical Journal}}
  \textbf{\bibinfo{volume}{27}}, \bibinfo{pages}{379--423}
  (\bibinfo{year}{1948}).

\bibitem{LEMPELZIV76}
\bibinfo{author}{Abraham, L.} \& \bibinfo{author}{Jacob, Z.}
\newblock \bibinfo{title}{On the complexity of finite sequences}.
\newblock \emph{\bibinfo{journal}{IEEE Trans. Inf. Theor.}}
  \textbf{\bibinfo{volume}{22}}, \bibinfo{pages}{75--81}
  (\bibinfo{year}{1976}).

\bibitem{Lesot:2009:SMB:1479242.1479248}
\bibinfo{author}{Lesot, M.~J.}, \bibinfo{author}{Rifqi, M.} \&
  \bibinfo{author}{Benhadda, H.}
\newblock \bibinfo{title}{Similarity measures for binary and numerical data: a
  survey}.
\newblock \emph{\bibinfo{journal}{Int. J. Knowl. Eng. Soft Data Paradigm.}}
  \textbf{\bibinfo{volume}{1}}, \bibinfo{pages}{63--84} (\bibinfo{year}{2009}).

\bibitem{Bishop:2006:PRM:1162264}
\bibinfo{author}{Bishop, C.~M.}
\newblock \emph{\bibinfo{title}{Pattern Recognition and Machine Learning
  (Information Science and Statistics)}} (\bibinfo{publisher}{Springer-Verlag
  New York, Inc.}, \bibinfo{address}{Secaucus, NJ, USA}, \bibinfo{year}{2006}).

\bibitem{10.1371/journal.pone.0094137}
\bibinfo{author}{Amancio, D.~R.} \emph{et~al.}
\newblock \bibinfo{title}{A systematic comparison of supervised classifiers}.
\newblock \emph{\bibinfo{journal}{PLoS ONE}} \textbf{\bibinfo{volume}{9}},
  \bibinfo{pages}{e94137} (\bibinfo{year}{2014}).

\bibitem{ebrahimpour2013automated}
\bibinfo{author}{Ebrahimpour, M.} \emph{et~al.}
\newblock \bibinfo{title}{Automated authorship attribution using advanced
  signal classification techniques}.
\newblock \emph{\bibinfo{journal}{PloS ONE}} \textbf{\bibinfo{volume}{8}},
  \bibinfo{pages}{e54998} (\bibinfo{year}{2013}).

\bibitem{amancio2015concentric}
\bibinfo{author}{Amancio, D.~R.}, \bibinfo{author}{Silva, F.~N.} \&
  \bibinfo{author}{Costa, L.~F.}
\newblock \bibinfo{title}{Concentric network symmetry grasps authors' styles in
  word adjacency networks}.
\newblock \emph{\bibinfo{journal}{EPL (Europhysics Letters)}}
  \textbf{\bibinfo{volume}{110}}, \bibinfo{pages}{68001}
  (\bibinfo{year}{2015}).

\bibitem{Navigli:2009:WSD:1459352.1459355}
\bibinfo{author}{Navigli, R.}
\newblock \bibinfo{title}{Word sense disambiguation: A survey}.
\newblock \emph{\bibinfo{journal}{ACM Comput. Surv.}}
  \textbf{\bibinfo{volume}{41}}, \bibinfo{pages}{10:1--10:69}
  (\bibinfo{year}{2009}).

\bibitem{dorogovtsev2002evolution}
\bibinfo{author}{Dorogovtsev, S.~N.} \& \bibinfo{author}{Mendes, J. F.~F.}
\newblock \bibinfo{title}{Evolution of networks}.
\newblock \emph{\bibinfo{journal}{Advances in physics}}
  \textbf{\bibinfo{volume}{51}}, \bibinfo{pages}{1079--1187}
  (\bibinfo{year}{2002}).

\end{thebibliography}

\end{document}